# Deep-Learning-Empowered Inverse Design for Freeform Reconfigurable Metasurfaces

Changhao Liu[1,2], Fan Yang[1,2,*], Maokun Li[1,2], Shenheng Xu[1,2]

[1]Department of Electronic Engineering, Tsinghua University, Beijing, 100084, China.
[2]Beijing National Research Center for Information Science and Technology (BNRist), Beijing, 100084, China.

*Corresponding author: fan_yang@tsinghua.edu.cn

## Abstract

The past decade has witnessed the advances of artificial intelligence with various applications in engineering. Recently, artificial neural network empowered inverse design for metasurfaces has been developed that can design on-demand meta-atoms with diverse shapes and high performance, where the design process based on artificial intelligence is fast and automatic. However, once the inverse-designed static meta-atom is fabricated, the function of the metasurface is fixed. Reconfigurable metasurfaces can realize dynamic functions, while applying artificial intelligence to design practical reconfigurable meta-atoms inversely has not been reported yet. Here, we present a deep-learning-empowered inverse design method for freeform reconfigurable metasurfaces, which can generate on-demand reconfigurable coding meta-atoms at self-defined frequency bands. To reduce the scale of dataset, a decoupling method of the reconfigurable meta-atom based on microwave network theory is proposed at first, which can convert the inverse design process for reconfigurable coding meta-atoms to the inverse design for static structures. A convolutional neural network model is trained to predict the responses of free-shaped meta-atoms, and the genetic algorithm is applied to generate the optimal structure patterns rapidly. As a demonstration of concept, several inverse-designed examples are generated with different self-defined spectrum responses in microwave band, and an inverse-designed wideband reconfigurable metasurface prototype is fabricated and measured for beam scanning applications with broad bandwidth. Our work paves the way for the fast and automatic design process of high-performance reconfigurable metasurfaces.

## Introduction

Metasurfaces, originating from 3D bulk metamaterials, are composed of artificial ultra-thin meta-atoms with periodic or quasi-periodic arrangement on a surface. In the past decade, metasurfaces emerge with unparalleled degree of freedom to manipulate electromagnetic (EM) wave in diverse functions, such as anomalous beam steering[1,2], invisibility cloaking[3,4], metalenses[5], and hologram[6,7]. The recent years have also witnessed the rapid development of reconfigurable metasurfaces, from analog reconfigurable metasurfaces[8,9], digital coding metasurfaces[10] to space-time coding metasurfaces[11]. By applying tunable materials to form meta-atoms or loading lumped tunable devices on the meta-atom structures, the reconfigurable metasurfaces can realize dynamic EM wave manipulating in real time. Lately, benefited from the great advances of artificial intelligence (AI), linking metasurfaces and deep learning techniques is a new research highlight nowadays[12], and various 'Meta + AI' applications are proposed, such as metasurface-enabled optical computation[13-15], AI-empowered waveform design[16], and intelligent metasurface imaging[17,18].

The idea of inverse design for metasurfaces based on artificial neural network (ANN) also follows the intelligent evolving trend of metasurfaces. To design metasurfaces, some conventional methods have been explored, but they have some shortcomings. Firstly, a direct theoretical relation between a given meta-atom structure and its EM responses is a long-standing challenge, which is not well-developed up to now. Therefore, numerical simulations for solving Maxwell's equations are necessary to design specific meta-atoms conventionally. However, the full-wave numerical algorithms involve matrix inversion (finite element method, FEM) or iterations (finite difference time domain method, FDTD)[19], which take much time before obtaining responses, leading to tremendous time cost in repeated simulations and iterative adjustments when designing meta-atoms. Besides, it is highly empirical to select a proper basic structure of the meta-atom to meet the performance targets, which requires long-term experience of designers, and even professional designers may not come up with the optimal structures yet. Accordingly, the conventional design methods lack generality, and the lengthy design process hinders the development of metasurfaces. In recent years, driven by big datasets and powerful computing platforms, AI methods are showing the enormous potential to find the inherent link between structures and responses. The reason is that there is an inner similarity between ANN and numerical simulation algorithms, where the map between meta-atom structure and responses is a complex nonlinear relationship, while the one of the main advantages of ANN is to deal with these nonlinear problems[20]. Therefore, forward ANN prediction models are employed to replace the numerical simulation and save the iteration time. Without simulation burden, intelligent method empowered inverse design for metasurfaces shows great advantages, such as fast design process, automatic and programmed procedure, large solution space, as well as high performance.

In early years, optimization methods, like genetic algorithm (GA), particle swarm optimization (PSO) and so on, are employed to design antennas and photonic devices inversely and automatically[21-26], where these algorithms can generate optimal structures at given response requirements by reducing the loss iteratively. Nevertheless, the forward map from structure to responses is enabled by numerical simulation, which is still time-costly and lacks generality. With the rise of AI from last decade on, deep neural networks (DNNs) are utilized to take the place of numerical simulation for forward prediction, and the global optimization methods are used for inverse design[27-32]. Once the neural network is trained, the inverse design process is fast and flexible. DNNs can learn how the parameter changes of fixed-form patterns affect the responses, but they cannot design patterns out of the pre-defined forms. To design freeform meta-atoms, the meta-atom is divided into pixels, and convolutional neural network (CNN) is proposed to make forward predictions from the pixel patterns to responses[33]. Besides, since the multiple meta-atoms may generate similar responses, direct inverse mapping from responses to structures using neural networks is an ill-posed problem, and the training process may not converge[20]. To tackle this problem, advanced generative network models are utilized for direct inverse design for freeform meta-atoms, such as generative adversarial networks (GANs)[34-37] and variational autoencoders (VAEs)[38]. These networks can also operate on smaller dataset and generate less fragmented geometries. To further reduce the scale of dataset, transfer learning methods are introduced to design metasurfaces inversely[39-41]. Smaller dataset, faster training process and higher prediction accuracy are always in demand for ANN-based inverse design for metasurfaces[20,26,42-44].

Apart from employing more advanced network to design static metasurfaces, the future trend of inverse design for metasurfaces also evolves toward dynamic functions. Although numerous inverse designs for metasurfaces are presented, we find the inverse designs for reconfigurable metasurfaces are rarely reported. Only recently, inverse-designed reconfigurable metasurfaces based on phase change materials (PCMs) like $Ge_2Sb_2Te_5$ (GST), $Ge_2Sb_2Se_4Te$ (GSST) and $VO_2$ are designed in infrared band with fixed patterns[45,46], and simulation results show the feasibility of inverse design for an active metasurface. However, achieving a practical inverse-designed freeform reconfigurable metasurface still remains a challenging problem. This mainly results from the fact that a reconfigurable meta-atom has multiple scattering states, so the dataset has to contain each state of the reconfigurable meta-atom, where the scale of dataset will be expanded by multiple times, and it is time-consuming to generate datasets and train the model.

In this work, a novel idea of freeform reconfigurable metasurface based on inverse design is presented. To address the challenge that the dataset is tremendously increasing for reconfigurable meta-atoms, we suggest a method to decouple the meta-atom into static structures and lumped switches based on microwave network equivalence, and the responses of 1-bit reconfigurable coding meta-atom at two states can be obtained by simulating the meta-atom structure for only one time. Based on our theory, the inverse design for two-state reconfigurable meta-atom is converted to inverse design for the static structure, and then the scale of dataset required for 1-bit reconfigurable meta-atom is reduced by half. To realize high-performance 1-bit phase coding, the requirements for static structure responses can be pre-calculated at given switch parameters. To demonstrate the idea, a CNN model is proposed to make forward predictions from structure to responses, and GA is employed for inverse meta-atom design. Freeform reconfigurable meta-atoms can be generated immediately at flexible on-demand operation bands like wideband, single-band and multi-band once the network is trained. As an experimental verification, a wideband freeform reconfigurable 1-bit coding metasurface is fabricated and measured, with an outstanding bandwidth over 20% at 5.8-GHz band, and dynamic beam-steering application is demonstrated with the wideband metasurface. Our work opens the front door for the on-demand fast design of high-performance reconfigurable metasurfaces, which could promote the development of metasurfaces in the intelligent era.

## Results

**Concepts and theoretical analysis**

The idea of inverse design for reconfigurable metasurfaces is illustrated in Fig. 1. A freeform reconfigurable reflective meta-atom with a lumped switch under spatial wave illumination (Fig. 1a) can be modeled as a two-port microwave network (Fig. 1b), where the static structure and lumped switch are decoupled, under the condition that the size of the lumped switch is much smaller than the size of the meta-atom. A 2×2 scattering matrix **S** is used to denote the parameters of the static structure, and the switch is loaded at port 2 of the network, denoted by impedance $Z_L$. Suppose the structure is magnet-free and reciprocal, the reflection coefficient from free space $\Gamma_1$ can be derived as

$$\Gamma_1 = S_{11} + \frac{S_{12}^2 \Gamma_L}{1-S_{22}} \quad (1)$$

where $S_{11}$, $S_{12}$ and $S_{22}$ are elements of matrix **S**, and $\Gamma_L$ is the reflection coefficient of the switch. The relation between $\Gamma_L$ and $Z_L$ is given by microwave theory

$$\Gamma_L = \frac{Z_L-\eta}{Z_L+\eta} \quad (2)$$

where η is the free space impedance. For 1-bit reconfigurable coding meta-atom, the switch has two states, characterized by 0 and 1, leading to two-state $\Gamma_1$ with 1-bit code.

Usually, the static structure has less ohmic loss than the switch. Hence, it is further supposed that the static structure is lossless, so **S** is a unitary matrix according to microwave theory[47]. Based on the unitary condition of **S**, $\Gamma_1$ can be deduced as the following form

$$\Gamma_1 = e^{j\theta_{11}} \frac{A_{22}-e^{j\theta_{22}}\Gamma_L}{1-A_{22}e^{j\theta_{22}}\Gamma_L} \quad (3)$$

In this equation, $\theta_{11}$ is the phase term of $S_{11}$, and $A_{22}$ and $\theta_{22}$ are amplitude term and phase term of $S_{22}$. Detailed derivations from Eq. (1) to Eq. (3) are presented in Supplementary Note 1. It is noted that for 1-bit coding meta-atoms, the constant phase term $e^{j\theta_{11}}$ cannot affect amplitude or phase difference of $\Gamma_1$, so the performance of dynamic metasurface will not be affected if omitting the term $e^{j\theta_{11}}$. Accordingly, Eq. (3) can be simplified as the following term

$$\Gamma_1 = \frac{A_{22} - e^{j\theta_{22}}\Gamma_L}{1 - A_{22}e^{j\theta_{22}}\Gamma_L} \qquad (4)$$

In Eq. (4), it is obvious that the reflection coefficient $\Gamma_1$ from free space is only influenced by the scattering parameter $S_{22}$ from the switch port. Therefore, $\Gamma_1$ can be directly calculated based on $S_{22}$ and given switch parameters. In this case, the excitation setups for simulation of reconfigurable meta-atom can be converted from Fig. 1b to Fig. 1c, where the power source is applied at the switch port to obtain reflection coefficient with $\Gamma_2 = S_{22}$. In Fig. 1c, there is no switch, so the simulated structure is static. Accordingly, designers can simulate the static structure for only one time to obtain the two-state reflection coefficients, which can reduce the number of simulations by half, and thus scale of dataset can be reduced by half.

Based on the decoupling understanding and the simulation setup in Fig. 1c, the inverse design process can be generalized in Fig. 1d, where the tunable switch and static structures are treated separately. Usually, a group of two-state switch parameters are given in advance, namely, $Z_L^0$ and $Z_L^1$, or $\Gamma_L^0$ and $\Gamma_L^1$. To obtain 1-bit phase coding with high performance, the reflection phase difference should be around 180°, and the reflection amplitudes should be close, which leads to $\Gamma_1^0 = -\Gamma_1^1$ in ideal case. Hence, the desired response requirements $S_{22}^r$ of static structure can be derived by solving the equation

$$\frac{A_{22} - e^{j\theta_{22}}\Gamma_L^1}{1 - A_{22}e^{j\theta_{22}}\Gamma_L^1} = -\frac{A_{22} - e^{j\theta_{22}}\Gamma_L^0}{1 - A_{22}e^{j\theta_{22}}\Gamma_L^0} \qquad (5)$$

Analytic methods or numerical methods can both be applied to solve Eq. (5), and the analytic derivation process is presented in Supplementary Note 2. It is obvious that $S_{22}^r$ is only related with switch parameters. To meet the response requirements $S_{22}^r$, a deep-learning based inverse design process for the static structure is shown in Fig. 1d. The freeform structure patterns are encoded by pixel as the input of pre-trained ANN, where responses can be forward predicted using the ANN in parallel. To reduce the loss between the predicted responses and the required responses, the optimization method GA is employed for generating the next offspring of patterns. When the loss is minimized and converged after iteration, the optimal structure patterns are generated with desired responses. Then the lumped switch is re-loaded on the structure to compose the final reconfigurable meta-atom, and beam steering applications can be realized based on the inverse-designed metasurface (Fig. 1e). It is noted that the desired response requirements $S_{22}^r$ can be set at self-defined frequency bands once the ANN is trained, so on-demand meta-atoms can be generated by using GA with diverse spectrum responses, such as wideband, single-frequency, and multi-frequency responses.

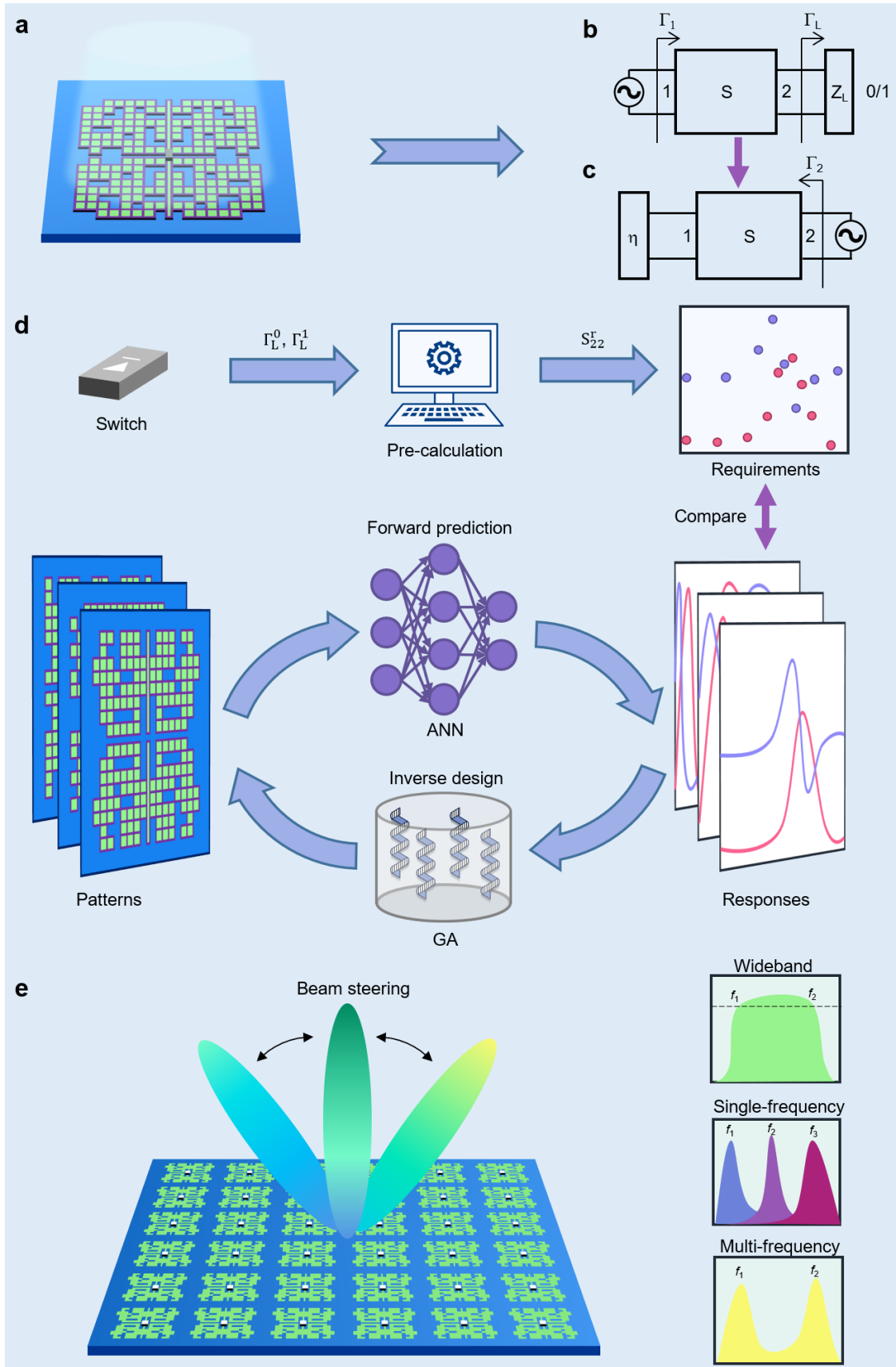

**Fig. 1 The concept of inverse design for reconfigurable metasurfaces. a** An illustration of a 1-bit reconfigurable reflective meta-atom, which is composed of the static structure and lumped switch. **b** The microwave network

equivalence of the reconfigurable meta-atom in (**a**), where the power is from free space, and reflection coefficient of the meta-atom is $\Gamma_1$. The reflection coefficient of the lumped switch is $\Gamma_L$, which has two coding states expressed as 0/1. **c** The transformed network model of (**b**) under lossless assumption, where the lumped power source is applied at the original location of lumped switch, and the reflection coefficient from the switch port is $\Gamma_2$. $\eta$ is the impedance of free space, i.e. 377 Ω. **d** The process of inverse design for reconfigurable metasurfaces. The required response target $S_{22}^r$ can be calculated based on the switch parameters $\Gamma_L^0$ and $\Gamma_L^1$ at 0/1 states. The freeform static structure patterns are inputs of the ANN for forward prediction of the responses, and GA is exploited to optimize the patterns to meet the response requirements, where the optimizing process is iterative. **e** The application of the inverse-designed reconfigurable metasurface for dynamic beam steering. Self-defined spectrum responses, such as wideband, single-frequency and multi-frequency, can be rapidly generated respectively based on inverse meta-atom design.

**Data generation**

As a proof of the concept, reconfigurable metasurfaces in microwave band are designed inversely in this work. Firstly, a home-made dataset is generated using Matlab-HFSS co-simulation (Fig. 2). Fig. 2a illustrates the pattern setups of the meta-atom for simulation. The central frequency is set at $f_0$=5.8 GHz, and thus the size of each meta-atom is $P = 25.8$ mm ($0.5\lambda_0$). A single-layer radio frequency (RF) dielectric substrate RF-35 is used with a thickness $h = 1.52$ mm, a dielectric constant $\varepsilon_0 = 3.5$, and a loss tangent $\tan\delta = 0.0018$. The front layer of the substrate is the freeform pixel pattern. To simplify the design, the meta-atom has 16×16 pixels, where 8×8 independent pixels compose a sub-atom, and the sub-atom is mirrored along two central lines to compose the full meta-atom for avoiding the cross-polarization conversion effect[34,40]. Each pixel has equal side length $L = 1.5$ mm, where '0' represents vacuum and '1' represents the metallic film. It is noted that there may exist single vertices at corners of the diagonally adjacent pixels as discussed in ref. 24, leading to the difficulty during fabrication, so the size of each pixel has already been enlarged by a little to create overlap at corners for avoiding the single vertices, where the spacing between each pixel is 1.375 mm in simulation and fabrication. For reconfigurable meta-atom setups, two design considerations should be noted. Firstly, the switch will be soldered on the center of the meta-atom, and to avoid the shorting of biasing circuit, two groups of sub-atoms are separated by a gap, where the gap length $gap = 0.3$ mm is the gap of the practical switch. Besides, to reserve the pad for soldering switches, two metallic leads are pre-placed on the substrate with width $W = 0.7$ mm and length $L_1 = 11.125$ mm, where the leads can also connect the sub-atoms and reserve space for biasing circuits. The substrate is backed by a reflective metallic ground plane. Based on our proposed theory, a lumped port is loaded on the original location of the switch to excite the meta-atom for simulation.

Based on the pattern setup, the 8×8 random pixels are generated by Matlab, where the meta-atom has $2^{8\times8} \approx 1.84\times10^{19}$ possible patterns. Besides, Matlab-HFSS co-simulation method is applied to generate the dataset automatically, where each sample contains 8×8 pixel codes and the simulated responses. The simulation frequency ranges from 2.8 GHz to 8.8 GHz with a step size of 0.1 GHz, so the response contains reflection amplitudes at 61 frequency points in real part and imaginary part respectively, where the amplitude responses range from -1 to 1. Finally, we collect 110 000 samples as the dataset, which are used to train and validate the network model. The detailed statistical analysis of dataset is further presented in Supplementary Note 3.

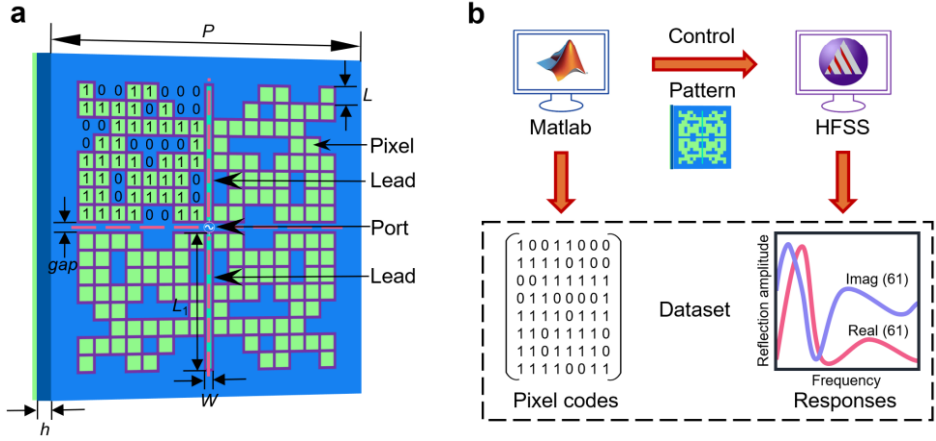

**Fig. 2 The basic setups of the meta-atom and generation process of the dataset. a** The setup of the freeform meta-atom. The meta-atom is coded by pixels with two-fold symmetry. Gap and leads are reserved for soldering switches, where the lumped port is applied at the center of the meta-atom to replace the lumped switch for simulation. **b** The dataset generation process, where Matlab generates random coding patterns and control HFSS software to simulate for the responses. A sample of dataset contains 8×8 pixel codes and reflection responses.

**CNN training and forward predictions**

To predict the responses at given structure patterns, we propose a deep-learning network model that is modified from the standard deep residual network (ResNet)[48], as shown in Fig. 3a. The proposed network has 73 layers, where the small kernel size k = 3 is used to reduce the scale of parameters. The residual block is a standard 2-layer basic block, where shortcut connection is introduced in each two layers. The activation function is ReLU, after convolution layer and batch normalization. The input of the network is the 16×16 pixel codes, and the output is the real and imaginary parts with length of 61 respectively. The network is trained on an Nvidia Tesla V100 GPU card with mini-batch size of 64, where 100 000 samples are used for training, and 10 000 samples are left for validation. Adam optimizer is used with initial learning rate of 0.001, which is divided by 10 when the loss curve meets plateau. To improve the robustness at abnormal values, mean absolute error (MAE) is employed as the loss function. The training process is converged after 152 epochs, which takes about 13 hours. The loss curves are shown in Fig 3b, where the training loss is $1.15 \times 10^{-2}$ and the validation loss is $2.87 \times 10^{-2}$ after convergence.

The simultaneous forward predictions of 1000 different patterns take within 0.25 seconds using a GPU card, which is improved by several orders of magnitude compared with conventional simulation process. To validate the prediction accuracy of the network intuitively, we randomly design two simple structures that are not in training or validation sets for simulation verification, which are shown in Fig. 3c, d. The predicted responses in real and imaginary parts are in good agreements with the simulation results, which means the deep network can learn the relations and make predictions from structure patterns to responses with acceptable error.

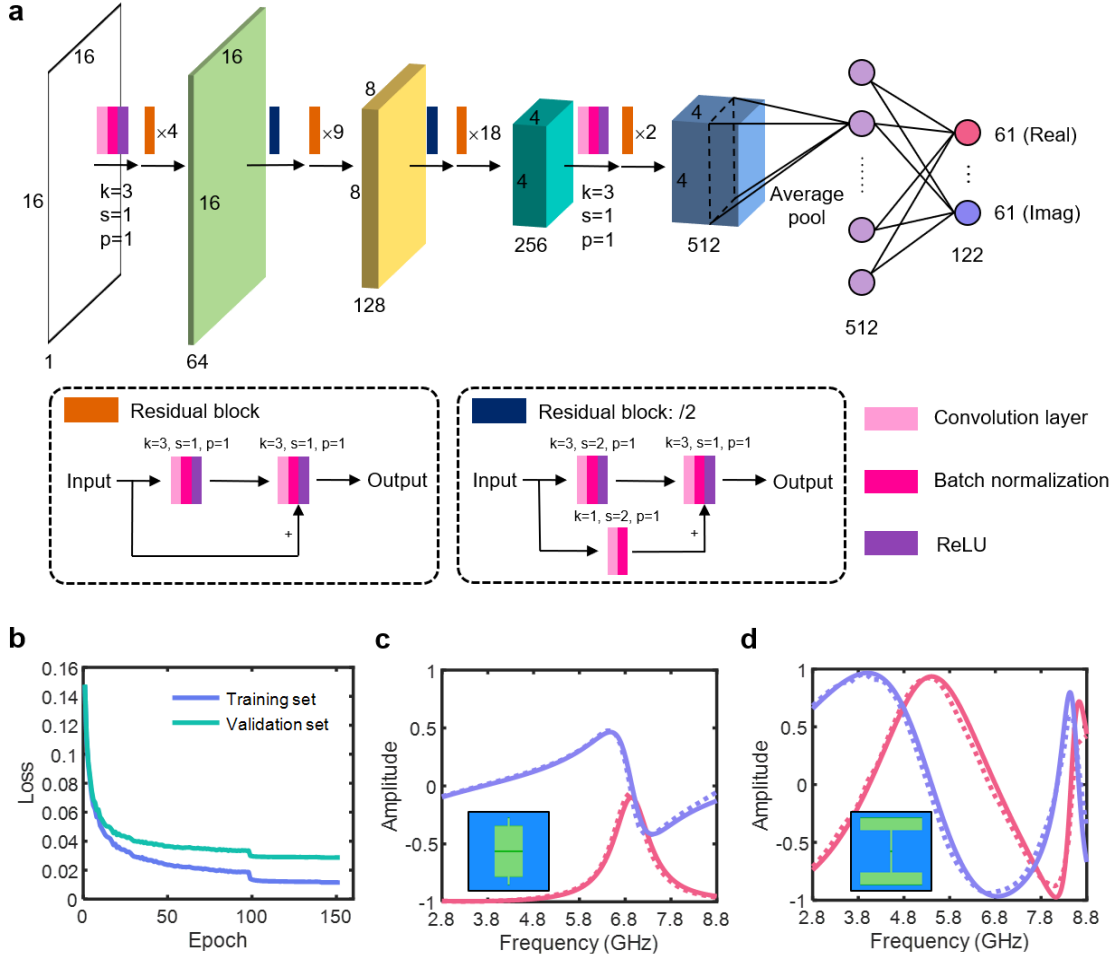

**Fig. 3 CNN model architecture and results. a** Schematic of the 73-layer ResNet model. k, s, p represent the kernel size, stride and padding of the convolution operations respectively. The input is the pattern of 16×16 pixel codes, and after convolution layers and full connection layers, the output is a 1×122 array, where the real part and the imaginary part own 61 values respectively. **b** The convergence curve of training and validation loss. **c, d** The forward prediction results of two random-designed simple meta-atoms. The prediction responses are depicted in dotted line, and the responses of simulation validation are depicted in solid line. Pink curves denote the real part and blue curves denote the imaginary part.

## Inverse design for on-demand reconfigurable meta-atoms

Based on the proposed theory, the required targets of the responses are pre-calculated based on switch parameters in advance. Here, the Skyworks SMP1340 PIN diode is used, which is modeled as lumped RLC series at 5.8 GHz: $R_{ON} = 1\ \Omega$, $L_{ON} = 450$ pH; $R_{OFF} = 10\ \Omega$, $L_{OFF} = 450$ pH, $C_{OFF} = 126$ fF. These equivalent parameters can be found in ref. 49. Solving Eq. (5), the required values of $S_{22}^r$ are $\mathrm{Re}(S_{22}^r) = -0.72$ for real part, and $\mathrm{Im}(S_{22}^r) = 0.01$ for imaginary part. Hence, the requirements should be set near the calculated values to realize high-performance 1-bit phase coding.

With ANN as a fast prediction machine to replace conventional simulation, diverse structures can be inversely designed by using global optimizers to find the optimal structures in a large solution space. Here, GA is utilized as the optimizer, with a population size of 1000, and the objective function of GA is defined by

$$\text{Fitness} = \frac{1}{\text{MAE}} \tag{6}$$

where MAE is the mean absolute error between the requirements and the predicted responses within the desired band. Standard selection, crossover and mutation processes are taken to generate offspring and to maximize the fitness. The detailed process of GA is shown in Supplementary Note 4. GA is run on the Nvidia Tesla V100 GPU card, and the iteration process stops after 300 generations, which takes 70 seconds to generate an optimal pattern.

Here, five examples with different spectrum responses are generated to demonstrate the proposed inverse design process, as shown in Fig. 4, where wideband (Fig. 4a), single-frequency (Fig. 4b, c), and multi-frequency (Fig. 4d, e) reconfigurable meta-atom patterns are generated respectively. For wideband meta-atom (Fig. 4a), the requirements are set near the calculated parameter $S_{22}^r$ from about 5.3 GHz to 6.3 GHz, while also considering the physical rules to improve the operation bandwidth. The prediction responses fit with the requirements and simulation results. When re-loading PIN diodes for full-wave simulation validations, the reflection amplitude loss is within 1.5 dB in desired band, and the phase difference is within 180°±45° from 5.35 GHz to 6.41 GHz, which correspond with the required frequency band. For single-frequency meta-atoms (Fig. 4b, c), the target central frequencies are set at 5.25 GHz and 6.35 GHz respectively, and the requirements, prediction results as well as simulation results are in good agreements. The performances are also validated by full-wave simulation, where the high-performance operation bands are 5.04 GHz – 5.57 GHz and 6.21 GHz – 6.68 GHz respectively, which accord with the target frequency bands. Furthermore, self-defined two-frequency operations can also be generated by inverse design, as shown in Fig. 4d, e. To avoid the possibility of wideband operation, the requirements are set to break the high-performance 1-bit phase at the intermediate band. For instance, to generate 1-bit phase coding at 5.15 GHz and 6.35 GHz in Fig. 4d, the out-of-band requirements with $\text{Re}(S_{22}^r) = -0.1$ and $\text{Im}(S_{22}^r) = 0$ are set at 5.75 GHz. Hence, the operation bands meet the requirements at 5.15 GHz and 6.35 GHz, while the phase difference is only 65.5° at 5.95 GHz, leading to poor out-of-band performance as desired. Apart from two-band operations at 5.15 GHz and 6.35 GHz, the optimal pattern at about 4.9 GHz and 6.6 GHz is also generated successfully in Fig. 4e. For more inverse-designed results, please refer to Supplementary Note 5. These examples have demonstrated that the inverse design process can rapidly generate on-demand reconfigurable meta-atoms of self-defined spectrum responses without violating the physical laws.

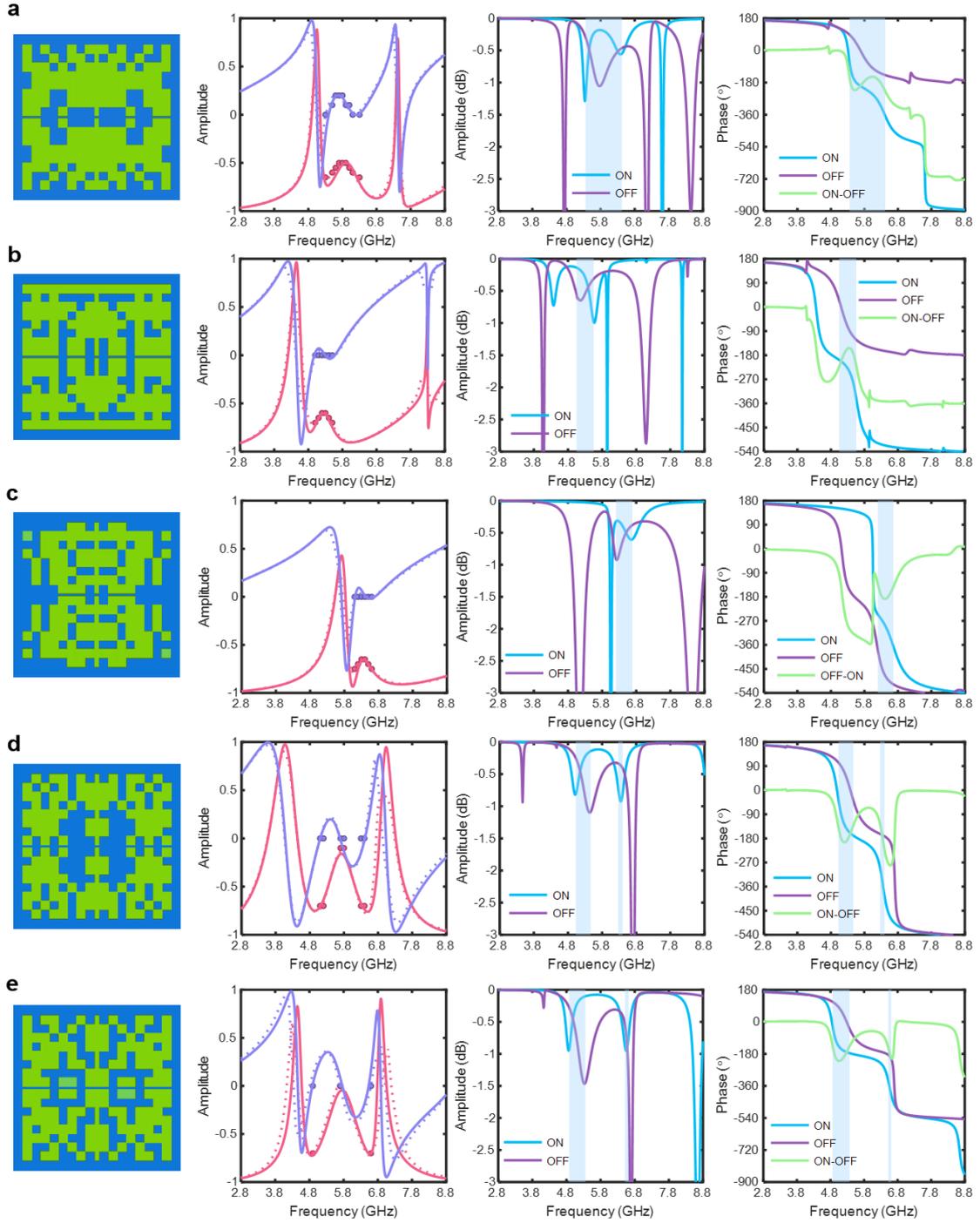

**Fig. 4 Results of inverse-designed patterns and simulation verifications.** (**a**) Wideband, (**b-c**) single-frequency, and (**d-e**) multi-frequency on-demand meta-atom patterns and results. The first column from left: inverse-generated patterns. The second column: the real part (pink) and imaginary part (blue) of the reflection coefficients from the lumped switch port, where the dots are requested values, dotted lines are forward predictions by the trained CNN model, and solid lines are simulation results. The third column and fourth column: simulated reflection amplitudes (third column) and phases (fourth column) from free space port with switches on the meta-atom, and blue areas denote the operation frequency band.

## Experimental verification

As an experimental verification of the inverse design method, a wideband reconfigurable metasurface composed of 16×16 meta-atoms in Fig. 4a is fabricated and measured. The prototype is fabricated using printed circuit board (PCB) technology, and one Skyworks SMP1340 PIN diode is soldered on the center of each meta-atom, as shown in Fig. 5a. Each reconfigurable meta-atom is controlled independently by a field programmable gate array (FPGA) board through biasing vias. The vias are placed at the null points of electric field at both 0/1 coding states to forbid the RF energy leakage (see Supplementary Note 6 for more information). Firstly, the reflection loss and phase differences are measured by far-field scattering test in a microwave anechoic chamber as shown in Fig. 5b. Results are shown in Fig. 5c, d, and the measured reflection amplitudes and phases accord with the simulation ones, where the 1-bit phase difference bandwidth is from 5.35 GHz to 6.42 GHz within 180°±45°. The losses of the reflection amplitudes are higher than simulation results, which is mainly due to the parasitic resistance when soldering the PIN diodes.

The pencil-beam forming and beam scanning applications are measured in a microwave anechoic chamber environment as shown in Fig. 5e. To generate the pencil-beam at $\vec{k} = (\theta_0, \phi_0)$ direction, the ideal phase compensation of the (m, n)th meta-atom is obtained in ref. 50

$$\varphi_{mn} = \varphi_{fmn} - \vec{k} \cdot \vec{r}_{nm} + \varphi_{ref} \tag{7}$$

where $\varphi_{fmn}$ is the phase delay from feed to (m, n)th meta-atom, $\varphi_{ref}$ is the reference phase constant, and $\vec{r}_{nm}$ is the position of the meta-atom. For 1-bit phase coding, the phase has two states, so the ideal phase is quantized to code 0 for $\varphi_{mn} \in [0, \pi)$ and code 1 for $\varphi_{mn} \in [\pi, 2\pi)$. The power intensity of the 0° pencil-beam is measured in spectrum domain at first, as shown in Fig. 5f. The power intensity is flat over the operation frequency from 5.4 GHz to 6.65 GHz with the 3-dB power bandwidth of 20.8%, which has the widest bandwidth at 5.8-GHz band (comparison of the wideband reconfigurable coding metasurfaces is shown in Supplementary Note 7). The measured 0° beam patterns in 2D plane from 5 GHz to 7 GHz are shown in Supplementary Note 8 to demonstrate the wideband beamforming performance. The beam scanning patterns at 30° and 60° in xoz and yoz planes are further measured at 5.8 GHz, as shown in Fig. 5g, h. The clear pencil-beam patterns are obtained at each desired angle directions, which demonstrates the reconfigurable metasurface can support beam scanning up to 60° in 2D full-space.

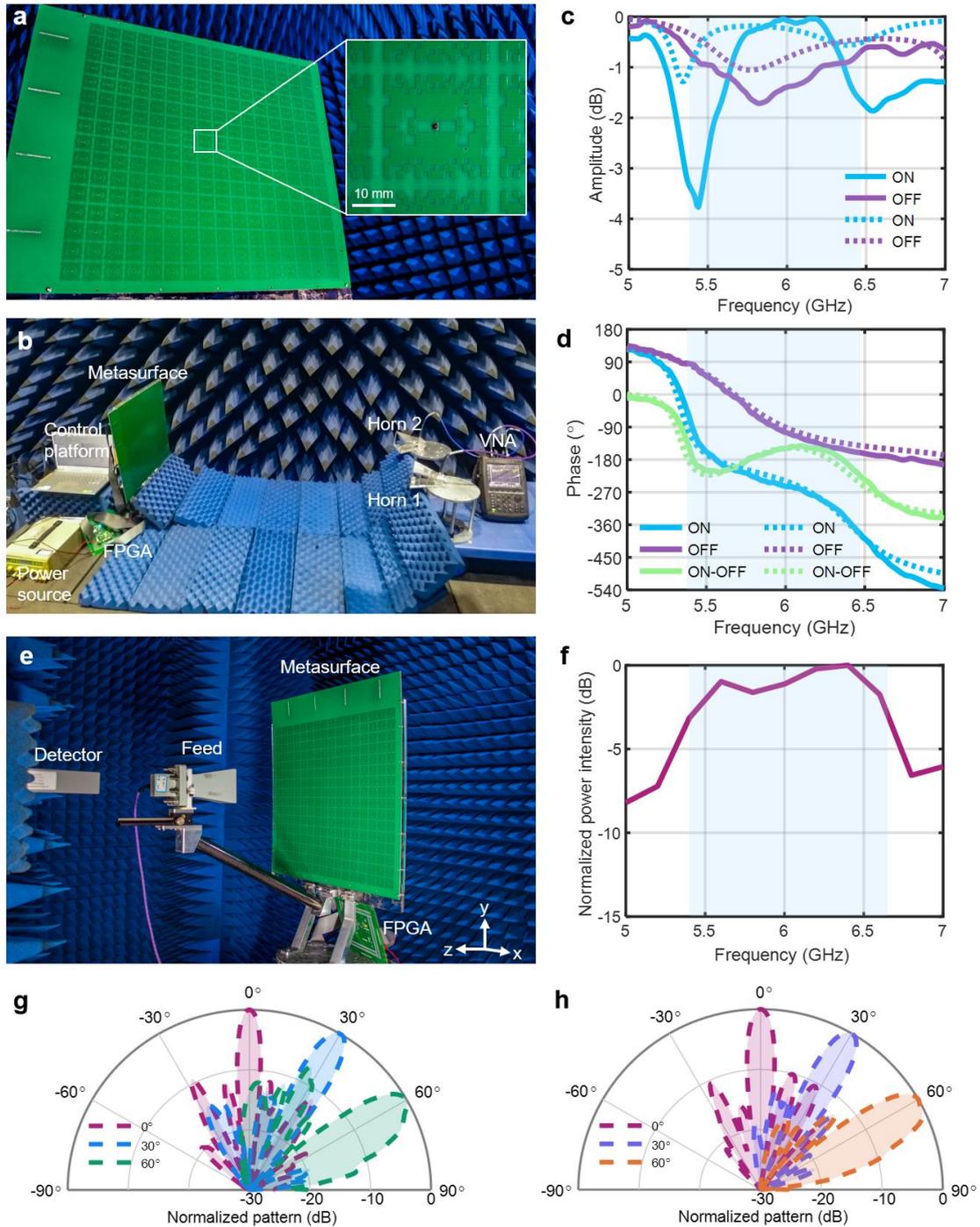

**Fig. 5 Experimental verifications of the inverse-designed wideband reconfigurable metasurface. a** Fabricated prototype of the wideband reconfigurable metasurface with 16×16 meta-atoms in Fig. 4a. PIN diodes are soldered on the center of the meta-atoms as tunable switches, and two vias are placed on one meta-atom for applying biasing voltage. The dark green region is metal and the light green region is the substrate, which are covered by solder mask. **b** Far-field scattering measurement environment for the reflection amplitude (**c**) and phase (**d**), where two horns are in mirror angles of the metasurface and are connected by a vector network analyzer (VNA). In (**c**) and (**d**), the dotted lines are simulated results, and the solid lines are measurement results. **e** Near-field beam scanning test environment. **f** The measured normalized power intensity of the metasurface with beam pointing at 0° direction. **g**, **h** Measured beam scanning patterns at 5.8 GHz in xoz plane and yoz plane respectively.

# Discussion

In conclusion, we proposed a deep-learning-empowered inverse design method for freeform reconfigurable metasurfaces. A novel method of decoupling reconfigurable meta-atoms is firstly introduced based on two-port network, which regards the reconfigurable meta-atom structure as a static structure, and thus only one simulation process is required to obtain reflection coefficients at multiple states. Based on the decoupling method, the tunable switch and static structure patterns are treated separately. The response requirements of structure patterns are pre-calculated analytically at given switch parameters. A CNN model is proposed and trained to make forward predictions from patterns to responses, and GA is employed to optimize the patterns iteratively for minimizing the loss between predicted responses and the pre-calculated requirements. As a proof of the concept, several on-demand inverse design examples are generated with self-defined spectrum responses in microwave band, and then a wideband reconfigurable metasurface is fabricated and measured to demonstrate the feasibility of the inverse design process for reconfigurable metasurfaces with exceptional performance, which is hard to realize via conventional design methods.

The inverse design method for freeform reconfigurable metasurfaces has unparalleled benefits over conventional methods, such as fast design process, flexible on-demand operation frequency bands, and large solution space for searching high-performance structures that is beyond the scope of conventional empirical structures. As an automatic design process, the inverse design method lowers the design threshold for promoting the development of reconfigurable metasurfaces. Furthermore, with inspiration by deep learning, the various inverse-generated high-performance meta-atom structures may offer a reference for researchers to explore the inherent physical relations between structures and EM responses.

# Materials and methods

**Full-wave simulation**

The numerical simulations for data generation and validations are conducted with Ansys Electronics Desktop HFSS 2018. The meta-atom is surrounded by periodic boundaries. Two sets of simulation setups are developed, where Floquet port is applied for simulation verifications with switches on the meta-atom as shown in Fig. 1b, while the lumped port excites at the original location of the switch to get $S_{22}$ in data generation and inverse design as shown in Fig. 1c. Besides, in data generation, Matlab 2020b is used to generate random patterns and control HFSS to simulate for reflection coefficients automatically.

**Model training and inverse design**

The model is developed under Python 3.7 and relies on Pytorch 1.1, which is trained on an Nvidia Tesla V100 GPU card. The training process takes about 13 hours. The optimization method GA is developed under Python 3.7. The forward predictions of 1000 patterns in parallel are accelerated by an Nvidia Tesla V100 GPU card. An Intel Xeon Gold 5118 CPU card is exploited for selection, crossover and mutation processes. The inverse design process takes about 70 seconds to generate an optimal on-demand pattern.

**Prototype fabrication**

As a demonstration of concept, a wideband prototype is fabricated using conventional printed circuit board (PCB) technology. A photograph of the prototype is shown in Fig. 5a. The metasurface is composed of 16×16 wideband meta-atoms in Fig. 4a with a spacing of 25.8 mm. The whole size of the prototype is 440 mm×510 mm, where the size of the meta-atom array is 412.8 mm×412.8 mm. The top substrate is RF-35 with a thickness of 1.52 mm, and the bottom substrate is FR4 with a thickness of 0.8 mm for placing biasing lines, and they are bonded together by a prepreg layer. The metallic ground is on the bottom side of RF-35 as a reflective ground plane. To control each meta-atom independently, four groups of 68 pins are soldered on the metasurface for electrical control, with 4 pins in each group connecting to the ground. Each pin is connected with each meta-atom through biasing lines. One Skyworks SMP1340 PIN diode is soldered on the center of one meta-atom, and 256 PIN diodes are soldered totally.

**Experimental setup and measurements**

The measurement setups are established in microwave anechoic chambers. The setup for far-field scattering measurements is shown in Fig. 5b, where two A-INFO LB-20180-NF horns are in 10° mirror angles to the normal line of the metasurface. The horns are connected by a Keysight N9951A vector network analyzer (VNA). The states of meta-atoms are programmed by the FPGA board with the control platform, where all states of the meta-atoms are switched between 0 and 1 to measure the meta-atom reflection performance. The results are calibrated by a metallic plate with equal size to the metasurface at the same location. The setup for near-field beamforming and beam scanning is shown in Fig. 5e, where an A-INFO LB-159-10-C-SF horn is employed as the feed. The feed is placed along the normal line of the metasurface, and the distance between the feed and the metasurface is 310 mm. A detector controlled by the stepping motor is placed behind the feed to scan in the xoy plane. The detector and the feed are connected by an Agilent E8361C VNA, and the amplitude and phase are recorded at each scanning position of the detector. Transform the near-field measured data, and the far-field beam patterns are obtained.

**Acknowledgements**

F.Y. acknowledges the support from the National Natural Science Foundation of China under Grant U2141233, as well as the support from THE XPLORER PRIZE.

applications (John Wiley & Sons, 2018).